%% file: mem_vs_pred.tex
\documentclass[pre,twocolumn,showpacs,superscriptaddress,preprintnumbers,floatfix]{revtex4-1}

\usepackage{etex}
\usepackage{ifpdf}
\usepackage{hyperref}
\usepackage{dcolumn}
\usepackage{url}
\usepackage{amsmath}
\usepackage{amscd}
\usepackage{amsfonts}
\usepackage{amssymb}
\usepackage{amsthm}
\usepackage{xfrac}
\usepackage{braket}
\usepackage{bm}   
\usepackage{bbm}
\usepackage{verbatim}
\usepackage{stmaryrd}
\usepackage{xcolor}
\usepackage{setspace}
\usepackage{tikz}
\usetikzlibrary{arrows}
\usetikzlibrary{automata}
\usetikzlibrary{positioning}
\usetikzlibrary{plotmarks}
\usepackage{pgfplots}
\pgfplotsset{compat=newest}

\tikzstyle{vaucanson}=[
  node distance=2.5cm,
  bend angle=15,
  auto,
  every loop/.style={},
  every edge/.style={->,draw=black,line width=1.2,>=latex,shorten <=1pt,shorten >=1pt},
  every state/.style={draw=black,line width=2,font=\large},
  loop right/.style={right,out=22,in=-22,loop},
  loop above/.style={above,out=112,in=68,loop},
  loop left/.style={left,out=202,in=158,loop},
  loop below/.style={below,out=292,in=248,loop},
  loop above right/.style={right,out=67,in=23,loop},
  loop above left/.style={left,out=157,in=113,loop},
  loop below left/.style={left,out=247,in=203,loop},
  loop below right/.style={right,out=337,in=293,loop},
]

\theoremstyle{plain}    
\theoremstyle{plain}    
\theoremstyle{plain}    
\theoremstyle{plain}    
\theoremstyle{plain}    
\theoremstyle{plain}    
\theoremstyle{plain}    
\theoremstyle{plain}    
\theoremstyle{plain}    
\theoremstyle{plain}    
\theoremstyle{plain}    
\theoremstyle{plain}    
\theoremstyle{plain}

\input{cmechabbrev}

\colorlet {R_color}    {blue}
\colorlet {k_color}    {black!30!green}

\def\clap#1{\hbox to 0pt{\hss#1\hss}}

\begin{document}

\title{The difference between memory and prediction in \\ linear recurrent networks}

\author{Sarah Marzen}
\email{semarzen@mit.edu}
\affiliation{Physics of Living Systems, Department of Physics, Massachusetts Institute of Technology,
Cambridge, MA 02139}

\date{\today}
\bibliographystyle{unsrt}

\begin{abstract}

Recurrent networks are trained to memorize their input better, often in the hopes that such training will increase the ability of the network to predict.  We show that networks designed to memorize input can be arbitrarily bad at prediction.  We also find, for several types of inputs, that one-node networks optimized for prediction are nearly at upper bounds on predictive capacity given by Wiener filters, and are roughly equivalent in performance to randomly generated five-node networks.  Our results suggest that maximizing memory capacity leads to very different networks than maximizing predictive capacity, and that optimizing recurrent weights can decrease reservoir size by half an order of magnitude.

\vspace{0.2in}

\end{abstract}

\preprint{arxiv.org:14XX.XXXX [physics.gen-ph]}

\maketitle


\setstretch{1.1}

\newcommand{\Abet}{\ProcessAlphabet}
\newcommand{\MS}{\MeasSymbol}
\newcommand{\ms}{\meassymbol}
\newcommand{\SSet}{\CausalStateSet}
\newcommand{\St}{\CausalState}
\newcommand{\st}{\causalstate}
\newcommand{\MxSt}{\AlternateState}
\newcommand{\MxSSet}{\AlternateStateSet}
\newcommand{\mxst}{\mu}
\newcommand{\mxstt}[1]{\mu_{#1}}
\newcommand{\StartMS}{\bra{\delta_p}}
\newcommand{\FSt}{\St^+}
\newcommand{\fst}{\st^+}
\newcommand{\PSt}{\St^-}
\newcommand{\pst}{\st^-}
\newcommand{\MT}{\mathcal{T}}
\newcommand{\mt}{\tau}
\newcommand{\ChanAlph}{\mathcal{Y}}


\vspace{0.2in}

Often, we remember for the sake of prediction.  Such is the case, it seems, in the field of echo state networks (ESNs) \cite{jaeger2001echo, jaeger2004harnessing}.  ESNs are large input-dependent recurrent networks in which a ``readout layer'' is trained to match a desired output signal from the present network state.  Sometimes, the desired output signal is the past or future of the input to the network.

If the recurrent networks are large enough, they should have enough information about the past of the input signal to reproduce a past input or predict a future input well, and only the readout layer need be trained.  Still, the weights and structure of the recurrent network can greatly affect the predictive capabilities of the recurrent network, and so many researchers are now interested in optimizing the network itself to maximize task performance \cite{lukovsevivcius2009reservoir}.

Much of the theory surrounding echo state networks centers on memorizing white noise, an input for which memory is essentially useless for prediction \footnote{Memory can be used to estimate the bias of the coin, but nothing else about the past provides a guide to the future input.}.  This leads to a rather practical question: how much of the theory surrounding optimal reservoirs, based on maximizing memory capacity \cite{jaeger2001short, white2004short, boedecker2012information, farkavs2016computational, baranvcok2014memory}, is misleading if the ultimate goal is to maximize predictive power?

We study the difference between optimizing for memory and optimizing for prediction in linear recurrent networks subject to scalar temporally-correlated input generated by countable Hidden Markov models.  Ref. \cite{hermans2010memory} gave closed-form expressions for memory function of continuous-time linear recurrent networks in terms of the autocorrelation function of the input, and closely studied the case of an exponential autocorrelation function.  Ref. \cite{goudarzi2016memory} gave similar expressions for discrete-time linear recurrent networks.  Ref. \cite{ganguli2008memory} gave closed-form expressions for the Fisher memory curve of discrete-time linear recurrent networks, which measure how much changes in input signal perturb the network state; for linear recurrent networks, this curve is independent of the particular input signal.

We differ from these previous efforts mostly in that we study both memory capacity and newly-defined ``predictive capacity''.  We derive an upper bound for predictive capacity via Wiener filters in terms of the autocorrelation function of the input.  Two surprising findings result.  First, predictive capacity is not typically maximized at the ``edge of criticality'', unlike memory capacity \cite{jaeger2001short, boedecker2012information, baranvcok2014memory}.  Instead, maximizing memory capacity can lead to minimization of predictive capacity.  Second, optimized one-node networks tend to achieve more than $99\%$ of the possible predictive capacity, while (unoptimized) linear random networks need at least five nodes to reliably achieve similar memory and predictive capacities, and ten-node nonlinear random networks cannot match the optimized one-node linear network.  The latter result suggests that optimizing reservoir weights can lead to at least half an order-of-magnitude reduction in the size of the reservoir with no loss in task performance.

\section{Model}

Let $s(n)$ denote the input signal at time $n$, and let $x(n)$ denote the network state at time $n$.  The network state updates as
\begin{equation}
x(n+1) = Wx(n) + s(n) v
\label{eq:setup}
\end{equation}
where $W,~v$ are two reservoir properties that we wish to optimize.  We restrict our attention to the case that $W$ is diagonalizable,
\begin{equation}
W = P \text{diag}(\vec{d}) P^{-1},
\label{eq:diagonalize}
\end{equation}
where $P$ is the matrix of eigenvectors of $W$ and $\vec{d}$ are the corresponding eigenvalues.
For reasons that will become clear later, we define a vector
\begin{equation}
\omega = P^{-1} v.
\end{equation}
We further assume that the input $s(t)$ has been generated by a countable Hidden Markov model, so that its autocorrelation function can be expressed as
\begin{equation}
R_{ss}(t) = \sum_{\lambda\in\Lambda} A(\lambda) \lambda^{|t|},
\label{eq:Rsst}
\end{equation}
where $\Lambda$ is a set of numbers with magnitude less than $1$.  See Ref. \cite{riechers2015pairwise} or the appendix.
To avoid normalization factors, we assert that
\begin{equation}
R_{ss}(0) = \sum_{\lambda\in\Lambda} A(\lambda) = 1.
\label{eq:varnorm}
\end{equation}
The power spectral density of this input process, with
\begin{equation}
R_{ss}(t) = \frac{1}{2\pi} \int_{-\pi}^{\pi} S(f) e^{ift} df,
\label{eq:WK}
\end{equation}
is
\begin{eqnarray}
S(f) &=& \sum_{k=\infty}^{\infty} R_{ss}(k) e^{-ifk} \\
&=& \sum_{\lambda\in\Lambda} A(\lambda) \sum_{k=-\infty}^{\infty} \lambda^{|k|} e^{-ifk} \\
&=& \sum_{\lambda\in\Lambda} A(\lambda) \frac{1-\lambda^2}{(1-\lambda e^{-if})(1-\lambda e^{if})}
\end{eqnarray}
by the Wiener-Khinchin theorem.

\section{Results}

The memory function is classically defined by \cite{jaeger2001short}
\begin{equation}
m(k) := p_k^{\top} C^{-1} p_k
\end{equation}
where
\begin{equation}
p_k = \langle s(n-k) x(n) \rangle_n
\end{equation}
and
\begin{equation}
C = \langle x(n) x(n)^{\top} \rangle_n.
\end{equation}
Due to Eq.~\ref{eq:varnorm}, we need not divide $p_k^{\top} C^{-1} p_k$ by the variance of the input.
This memory function is also the squared correlation coefficient between our remember or forecast of input $s(n-k)$ from network state $x(n)$ and the true input $s(n-k)$.

Memory capacity is usually defined as $\sum_{k=0} m(k)$, but since Eq.~\ref{eq:setup} updates $x(n)$ with $s(n-1)$ instead of $s(n)$, we have
\begin{equation}
MC = \sum_{k=1}^{\infty} m(k),
\end{equation}
and we define the \emph{predictive capacity} as
\begin{equation}
PC := \sum_{k=0}^{\infty} m(-k).
\end{equation}
Intuitively, $MC$ is higher when the present network state is better able to remember inputs, while $PC$ is higher when the present network state is better able to forecast inputs based on what it remembers of input pasts.

We have made an effort here to find the most useful expressions for $MC$ and $PC$, so that one might consider using the expressions here to calculate $MC,~PC$ instead of simulating the input and recurrent network.  As shown in the appendix,
\begin{equation}
PC = 2\pi \omega^{\top}\left(D_{PC} \odot B^{-1}\right) \omega.
\label{eq:PC}
\end{equation}
where
\begin{equation}
B := \int_{-\pi}^{\pi} S(f) \left(\frac{\omega}{e^{-if}-\vec{d}}\right) \left(\frac{\omega}{e^{if}-\vec{d}}\right)^{\top}df,
\end{equation}
which is related to $2\pi C$ by a similarity transform, and where
\begin{equation}
D_{PC} := \sum_{\lambda,\lambda'\in\Lambda} \frac{A(\lambda) A(\lambda')}{1-\lambda\lambda'} \left(\frac{1}{\lambda^{-1}-\vec{d}}\right)\left(\frac{1}{(\lambda')^{-1}-\vec{d}}\right)^{\top}.
\end{equation}
The expression for memory capacity is more involved:
\begin{eqnarray}
MC &=& 2\pi \omega^{\top}\left(D_{MC} \odot B^{-1}\right) \omega,
\label{eq:MC}
\end{eqnarray}
where the matrix $D_{MC}$ has entries
\begin{widetext}
\begin{eqnarray}
\left(D_{MC}\right)_{ij} &=& \sum_{\lambda,\lambda'\in\Lambda} A(\lambda) A(\lambda') \Big(\frac{1+d_i d_j \lambda (\lambda')^3+d_id_j \lambda' \lambda^3+d_id_j (\lambda\lambda')^{2}-d_id_j \lambda\lambda'-d_id_j (\lambda')^{2}-d_i d_j \lambda^{2}-d_i^2d_j^2}{(1-\lambda\lambda')(1-d_i\lambda)(1-d_i\lambda')(1-d_j\lambda)(1-d_j\lambda')(1-d_i d_j)} \nonumber \\
&& -\frac{d_i(1-d_id_j)\lambda^{2}\lambda'+d_j (1-d_id_j) \lambda(\lambda')^{2}}{(1-\lambda\lambda')(1-d_i\lambda)(1-d_i\lambda')(1-d_j\lambda)(1-d_j\lambda')(1-d_i d_j)}\Big)
\end{eqnarray}
\end{widetext}
as shown in the appendix.  Together, these expressions explain why simple linear ESNs \cite{fette2005short} can perform just as well as non-simple linear ESNs on maximization of $MC$; from Eq.~\ref{eq:MC}, the memory capacity of a linear ESN is the same as the memory capacity of a simple linear ESN with $v= \omega$ and $W=\text{diag}(\vec{d})$.

\begin{figure}
\includegraphics[width=0.45\textwidth]{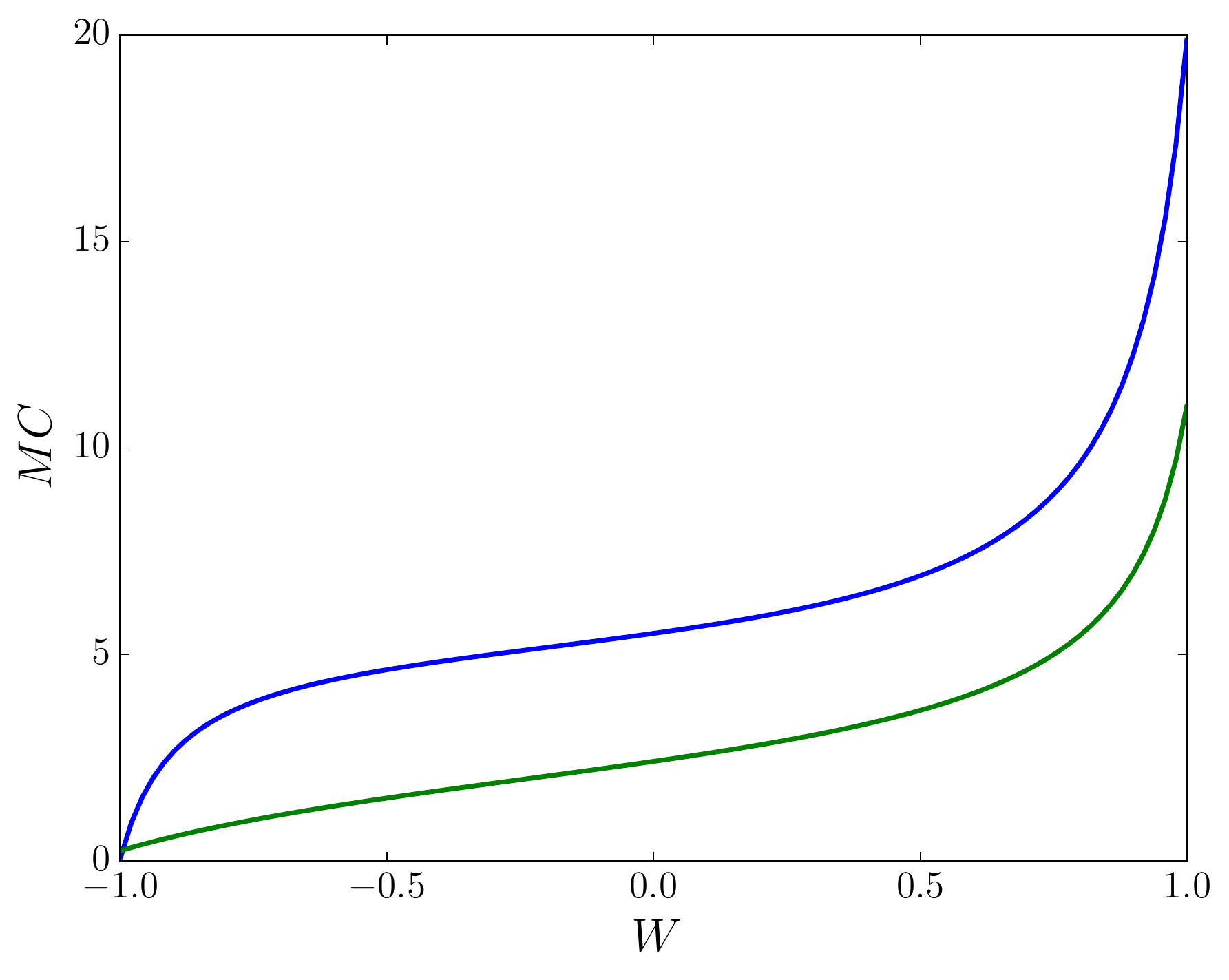}
\includegraphics[width=0.45\textwidth]{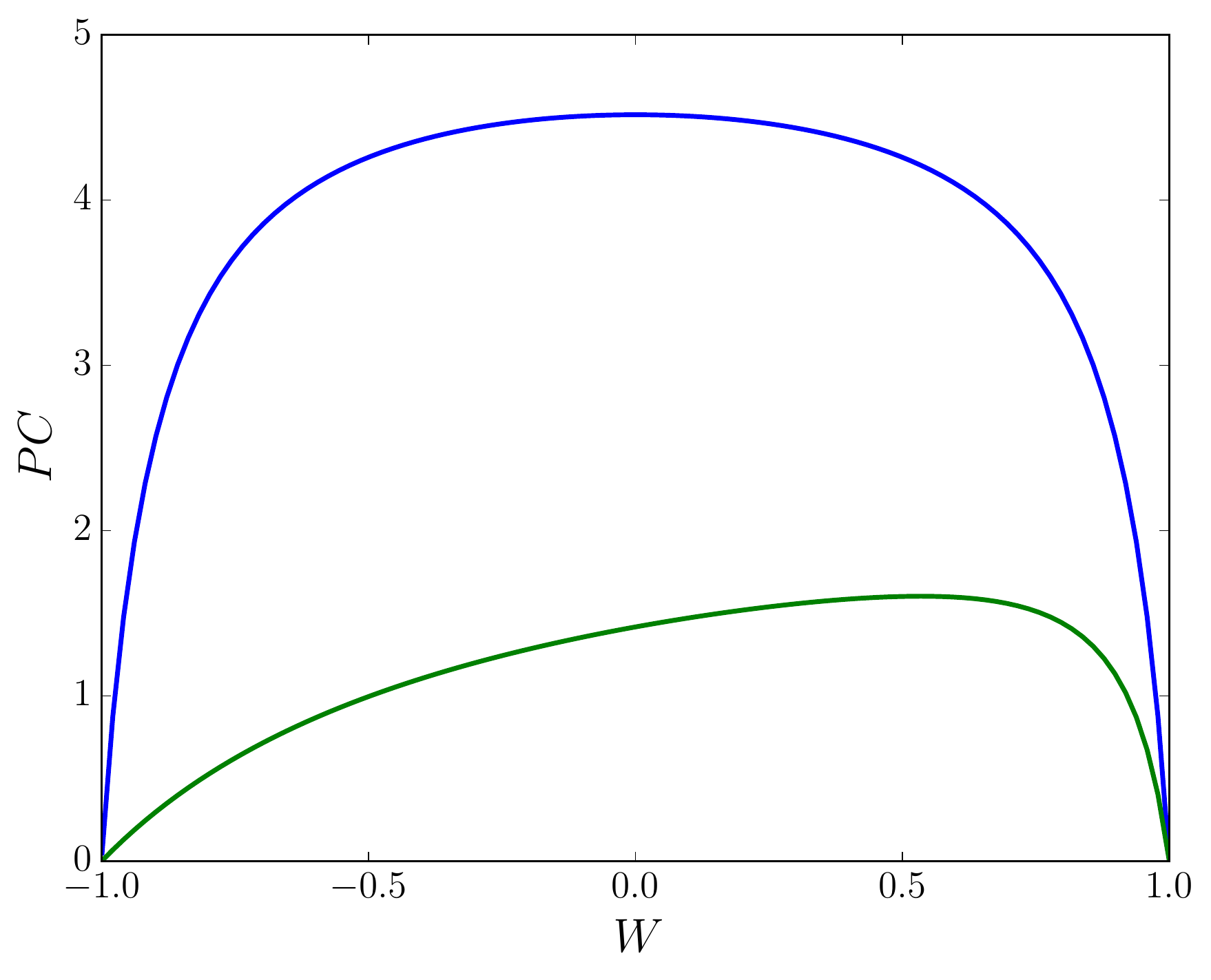}
\caption{$MC$ (top) and $PC$ (bottom) as a function of $W$ for $R_{ss}(t) = e^{-0.1|t|}$ (blue) and $R_{ss}(t) = \frac{1}{2} e^{-0.1|t|}+\frac{1}{2}e^{-|t|}$ (green), computed using Eqs.~\ref{eq:PC} and \ref{eq:MC} in the main text.  While $PC$ is maximized for some intermediate $W$ that depends on the input signal, $MC$ is maximized in the limit $W\rightarrow 1$.  When $|W|\geq 1$, the network no longer satisfies the echo state property, and so we only calculate $PC,~MC$ for $|W|<1$.}
\label{fig:easy}
\end{figure}

\begin{figure}
\includegraphics[width=0.45\textwidth]{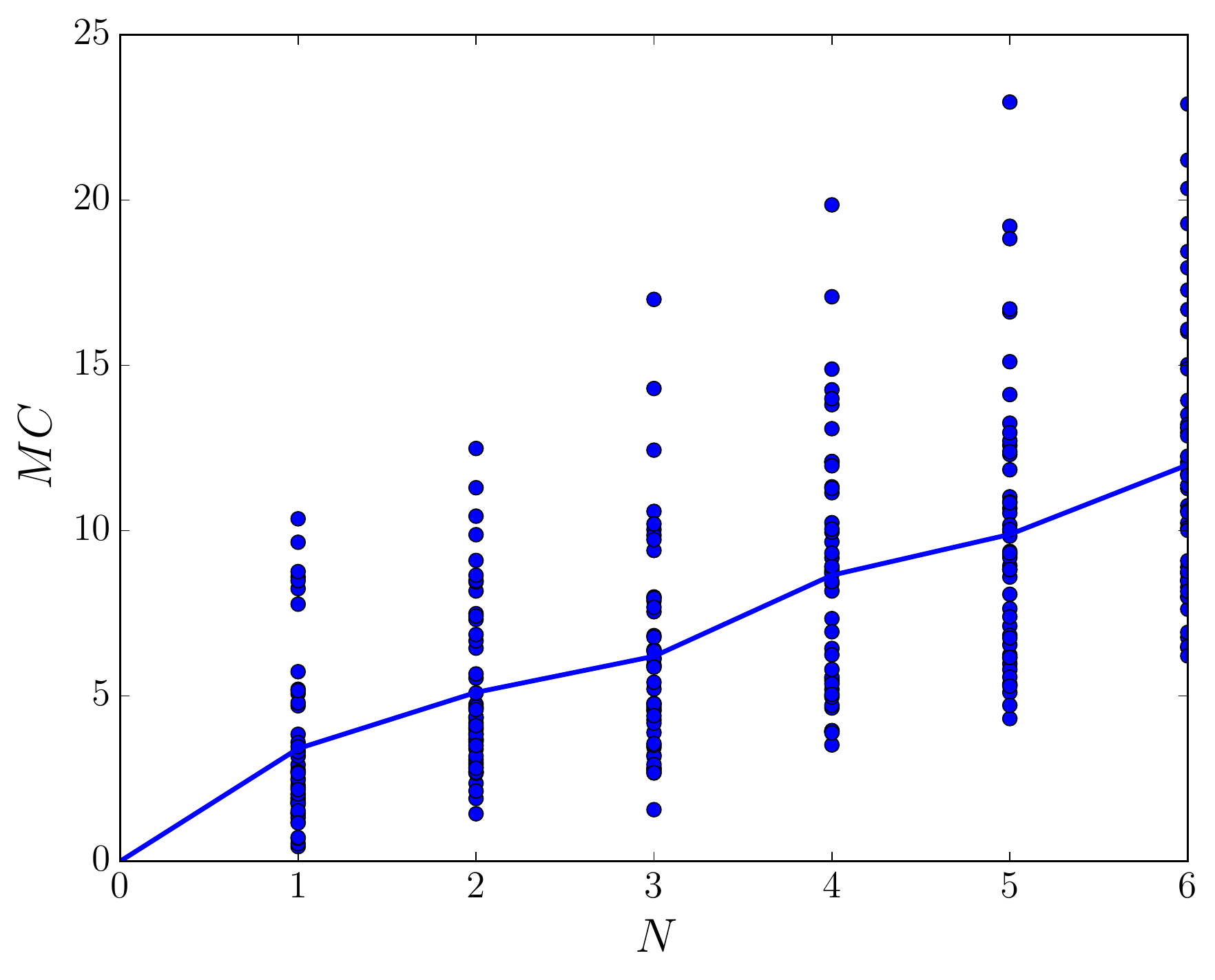}
\includegraphics[width=0.45\textwidth]{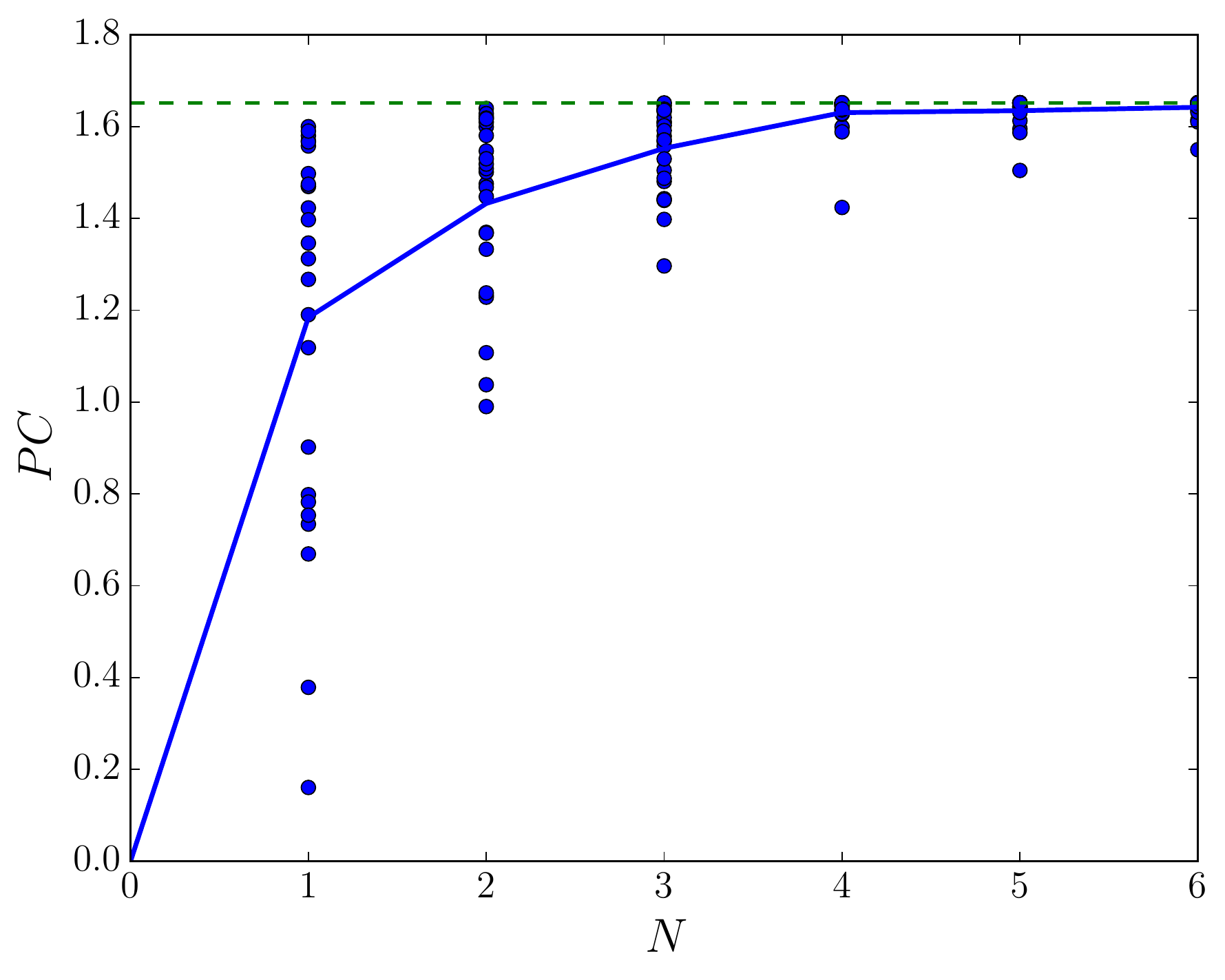}
\caption{$MC$ (top) and $PC$ (bottom) as a function of $N$ for $R_{ss}(t) = \frac{1}{2} e^{-0.1|t|}+\frac{1}{2}e^{-|t|}$ and $\omega,~\vec{d}$ drawn randomly: $\omega_i \sim \mathcal{U}[0,1],~d_i\sim\mathcal{U}[-1,1]$.  The signal-limited maximal $PC$ is shown in green, whereas $MC$ is network-limited.  Both $MC,~PC$ were computed using Eqs.~\ref{eq:PC} and \ref{eq:MC} in the main text.}
\label{fig:random}
\end{figure}

It is unsurprising but not often mentioned that the reservoirs which maximize memory capacity are different than the reservoirs that maximize predictive capacity.  To illustrate how different the two reservoirs might be, we consider the capacity of a one-node network subject to two types of input.

The first type of input considered has autocorrelation $R_{ss}(t) = e^{-\alpha|t|}$.  Some algebra reveals that $MC = \frac{e^{4\alpha}-2e^{\alpha}W+2e^{3\alpha}W-W^2}{(e^{2\alpha}-1)(e^{2\alpha}-W^2)}$ and $PC = \frac{e^{2\alpha}(1-W^2)}{(e^{2\alpha}-1)(e^{2\alpha}-W^2)}$.  Inspection of these formulae or inspection of the plots of these formulae in Fig.~\ref{fig:easy} (blue lines) for $\alpha=0.1$ shows that $MC$ is maximized at the ``edge of criticality'', $W\rightarrow 1$, at which point, $x(n)$ is an average of observed $s(n)$-- i.e., $x(n) = \langle s(k)\rangle_{k\leq n}$.  
Interestingly, at that point, $PC$ is minimized, i.e. $PC=0$.  Instead, for this particular input, $PC$ is maximized at $W=0$, at which point, $x(n) = s(n-1)$-- i.e., $x(n)$ is the last observed input symbol.

Both memory and predictive capacity can increase without bound by increasing the length of temporal correlations in this input: $\lim_{W\rightarrow 1} MC = \coth (e^{\alpha/2})$, and $\lim_{W\rightarrow 0} PC = \frac{1}{2}(\coth\alpha-1)$.  These results mirror what was found in Ref. \cite{hermans2010memory} for continuous-time networks: $\lim_{W\rightarrow 1} MC= \frac{2}{\alpha}$ plus corrections of $O(\alpha)$, and $\lim_{W\rightarrow 0} PC = \frac{1}{2\alpha}$ plus corrections of $O(1)$.

It is a little strange to say that $W=0$ can maximize the predictive capacity of a reservoir, as $W=0$ implies that there essentially is no reservoir.  But such $\arg\max_W PC$ is unusual.  Consider input with $R_{ss}(t) = \frac{1}{2} e^{-0.1|t|}+\frac{1}{2}e^{-|t|}$ to a one-node network.  Memory capacity is still maximized as $W\rightarrow 1$, but predictive capacity is now maximized at $W \approx 0.8$.  See Fig.~\ref{fig:easy}(green).
Interestingly, we still minimize any error in memorization of previous inputs by storing (and implicitly guessing) their average value.

The scaling of capacity with the network size is also very different for memory and prediction.  Memory capacity $MC$ for linear recurrent networks famously scales linearly with the number of of nodes for linear recurrent networks \cite{jaeger2001short}.  Unlike memory, predictive capacity $PC$ is bounded by the signal itself.  The Wiener filter $k_{\tau}(n)$ minimizes the mean-squared error $\langle (s(n+\tau)-\hat{s}(n+\tau))^2 \rangle_n$ of future input $s(n+\tau)$ and a forecast of future input from past input, $\hat{s}(n+\tau) := \sum_{m=0}^{\infty} k_{\tau}(m) s(n-m)$.  Recall that minimizing mean-squared error is equivalent to maximizing the correlation coefficient between future input and forecast of this future input.  Hence, we can place an upper bound on predictive capacity $PC$ in terms of Wiener filters, which after some straightforward simplification shown in the appendix takes the form
\begin{equation}
PC \leq \sum_{\tau=0}^{\infty} \vec{r}_{\tau}^{\top} R^{-1} \vec{r}_{\tau}
\label{eq:PCBound1}
\end{equation}
where $(\vec{r}_{\tau})_i = R_{ss}(\tau+i)$ and $R_{ij} = R_{ss}(i-j)$.


\begin{figure}
\includegraphics[width=0.45\textwidth]{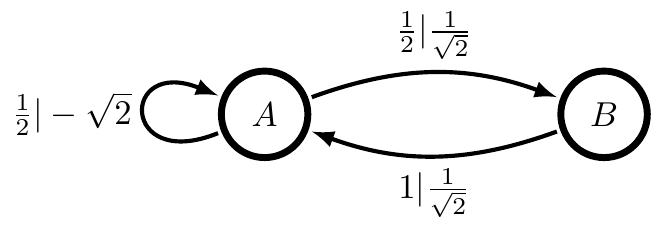}
\includegraphics[width=0.5\textwidth]{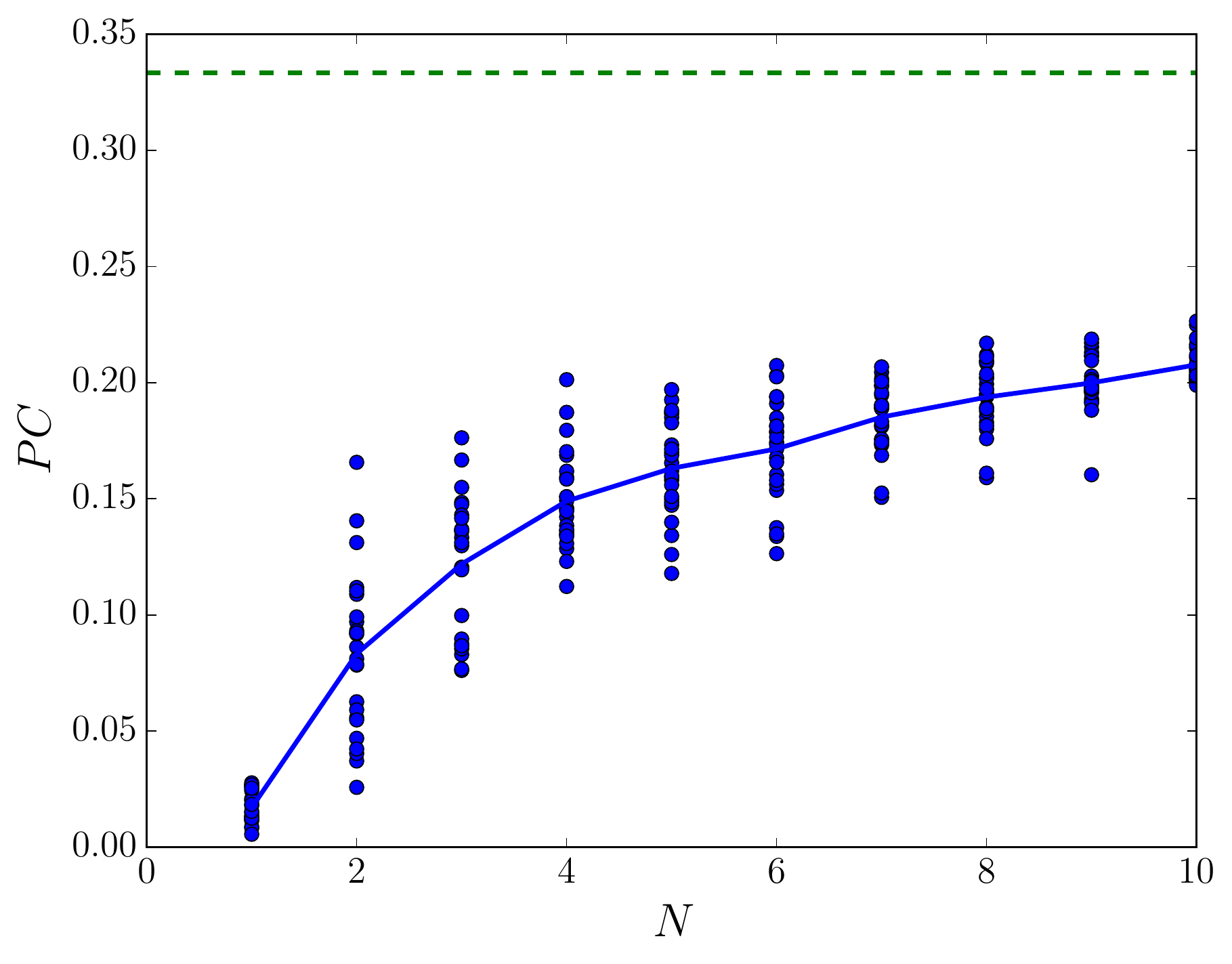}
\caption{At top, the Hidden Markov model generating input to the nonlinear recurrent network.  Edges are labeled $p(\ms|g)|\ms$, where $\ms$ is the emitted symbol and $p(\ms|g)$ is the probability of emitting that symbol when in hidden state $g$, and arrows indicate which hidden state one goes to after emitting a particular symbol from the previous hidden state.  This Hidden Markov model generates a zero-mean, unit-variance Even Process, which has autocorrelation function $R_{ss}(t) = \left(-\frac{1}{2}\right)^{|t|}$.  At bottom, the predictive capacity of random nonlinear recurrent networks whose evolution is given by Eq.~\ref{eq:nonlinearRNN} with $f(x)=\tanh (x)$ and entries of $W$ and $v$ drawn randomly: $W_{ij},~v_i \sim \mathcal{U}[0,1]$, where $W$ is then scaled so that its largest magnitude eigenvalue has absolute value $1/1.1$.  $25$ random networks are surveyed at each $N$, and the blue line tracks the mean.  The green line shows both the predictive capacity of the optimized one-node linear network and the upper bound from Eq.~\ref{eq:PCBound1}.}
\label{fig:nonlinearRNN}
\end{figure}

As $PC$ is at most finite, the scaling of $PC$ with the number of nodes of the network $N$ must be eventually $o(1)$.  See Fig.~\ref{fig:random}(bottom).  For instance, for $R_{ss}(t) = \frac{1}{2} e^{-0.1|t|}+\frac{1}{2}e^{-|t|}$, Eq.~\ref{eq:PCBound1} gives $PC \leq 1.652$, which is nearly attained by the optimal one-node network, for which $\max_W PC$ is $\approx 1.65$.  And this is not a special property for a cherry-picked input signal; similar results hold for other different, randomly chosen $\Lambda,~A_{\lambda}$ combinations not shown here.

The surprisingly good performance of optimized one-node networks leads us then to ask how big random (unoptimized) networks need be in order to achieve similar results.  Unoptimized random networks need $\approx 5$ nodes to reliably achieve similar results to the optimized one-node network for both memory and predictive capacity.  See Fig.~\ref{fig:random} in comparison to Fig.~\ref{fig:easy}.

Finally, we ask whether any of the lessons learned here for linear recurrent networks extend to nonlinear recurrent networks in which
\begin{equation}
x(n+1) = f(Wx(n)+s(n) v)
\label{eq:nonlinearRNN}
\end{equation}
for some nonlinear function $f$.  From Eq.~\ref{eq:xofn}, we see that linear recurrent networks forecast input via a linear combination of past input; therefore, as noted previously, their performance is bounded from above by the performance of Wiener filters.
The performance of nonlinear recurrent networks is bounded above by a quantity that depends on the nonlinearity, which in principle might surpass the bound on predictive capacity given by Eq.~\ref{eq:PCBound1}.

However, optimizing the weights of nonlinear recurrent networks is far more difficult than for linear recurrent networks.  This is illustrated by Fig. \ref{fig:nonlinearRNN}(bottom), which shows $\widehat{PC}$ of random nonlinear networks.  We estimate the predictive capacity from simulations via $\sum_{k=0}^{M} \hat{p}_k^{\top} \hat{C}^{-1} \hat{p}_k$, where $\hat{p}_k^{\top}$ is the sample covariance of $s(n+\tau)$ and $x(n)$, $\hat{C}$ is the sample variance of $x(n)$, and $M$ is taken to be $100$, as the correlation coefficient dies off relatively quickly.  The reservoir properties $W$ and $v$ are chosen randomly, in that both matrix elements $W_{ij}$ and vector elements $v_i$ are drawn randomly at uniform from the unit interval, and the matrix $W$ is rescaled so that the eigenvalue of maximum magnitude has magnitude of $1/1.1$, and the nonlinearity is set to $f(x) = \tanh x$.  The input to the network is generated by the Hidden Markov model shown in Fig.~\ref{fig:nonlinearRNN}(top).  For comparison, the green line shows the upper bound on predictive capacity for linear recurrent networks given by Eq.~\ref{eq:PCBound1}, which is achieved by one-node linear networks with $W=0$.  These numerical results are qualitatively similar to results attained when comparing the memory capacity of linear and nonlinear recurrent networks, in that linear networks tend to outperform nonlinear networks \cite{ganguli2008memory, toyoizumi2012nearly}.

\section{Discussion}

The famous Wiener filter is a linear combination of the past input signal that minimizes the mean-squared error between said linear combination and a future input.  Linear recurrent networks are, in some sense, an attempt to approximate the Wiener filter under constraints on the kernel that come from the structure of the recurrent network.  Here, the linear filter is not allowed access to all the past of the signal, but is only allowed access to the echoes of the signal past provided by the present state of the nodes.  The advantage of such an approximation is that one only need store the present network state, as opposed to storing the entire past of the input signal.  In other words, the present network state provides a nearly sufficient ``echo'' of the input signal's past for input prediction.

We have studied the resource savings that can come from optimizing the recurrent network and readout weights, as opposed to just optimizing the readout weights.  Surprisingly, we find that a network designed to maximize memory capacity has arbitrarily low predictive capacity; see Fig.~\ref{fig:easy}.  More encouragingly, we find that an optimized single-node linear recurrent network is essentially equivalent in terms of both memory and predictive capacity to a five-node random linear recurrent network, and near maximal predictive capacity.  Finally, numerical results suggest that nonlinear recurrent networks have more difficulty achieving high predictive capacity relative to the Wiener filter-placed upper bound, even though these nonlinear networks might in principle surpass such an upper bound.

It is unclear whether or not the factor-of-five will generalize to nonlinear recurrent networks or for inputs generated by uncountable Hidden Markov models, e.g. the output of chaotic dynamical systems.  Perhaps more importantly, predictive capacity is not necessarily the quantity that we would most like to maximize \cite{collins2016capacity}.  Hopefully, the differences between memory and predictive capacity presented here will stimulate the search for more task-appropriate objective functions and for more reservoir optimization recipes.

\acknowledgments

We owe substantial intellectual debts to A Goudarzi, J P Crutchfield, S Still and A Bell and additionally thank I Nemenman, N Ay, C Hillar, S Dedeo and W Bialek for very useful conversations.
S.M. was funded by an MIT Physics of Living Systems Fellowship.


\section*{References}
\bibliography{chaos}

\appendix

\begin{widetext}

\section{Autocorrelation function of Hidden Markov models}

This is a simple version of the argument in Ref. \cite{riechers2015pairwise} that assumes diagonalizability of the transition matrix.  Let $T^{(\ms)}$ be the labeled transition matrices of the Hidden Markov model, let
\begin{equation}
T = \sum_{\ms} T^{(\ms)}
\end{equation}
be the transition matrix, and let $\vec{p}_{eq} = \text{eig}_1(T)$ be the stationary distribution over the hidden states.  Assuming zero-mean input, we have
\begin{eqnarray}
R(t) &=& \langle x(t-1) x(0) \rangle \\
&=& \sum_{\ms,\ms'} \ms \ms' \Prob(\MS_{t-1}=\ms,\MS_0=\ms') \\
&=& \sum_{\ms,\ms'} \ms \ms'  \vec{1}^{\top} T^{(\ms)} T^t T^{(\ms')} \vec{p}_{eq} \\
&=& \sum_{\ms,\ms'} \vec{1}^{\top} \left(\ms T^{(\ms)}\right) T^{t} \left(\ms' T^{(\ms')}\right) \vec{p}_{eq} \\
&=& \vec{1}^{\top} \left(\sum_{\ms} \ms T^{(\ms)}\right) T^t \left(\sum_{\ms} \ms T^{(\ms)}\right) \vec{p}_{eq}.
\end{eqnarray}
If $T$ is diagonalizable (and it typically is), then $T = P\text{diag}(\vec{\lambda})P^{-1}$ leads to
\begin{eqnarray}
R(t) &=& \vec{1}^{\top} \left(\sum_{\ms} \ms T^{(\ms)}\right) P \text{diag}(\vec{\lambda}^t) P^{-1} \left(\sum_{\ms} \ms T^{(\ms)}\right) \vec{p}_{eq},
\end{eqnarray}
and so $R(t)$ is a linear combination of $\lambda_i^t$.

\section{Derivation of closed-form expressions for $PC,~MC$}

From Eq.~\ref{eq:setup}, we have
\begin{equation}
x(n) = \left(\sum_{k=1}^{\infty} W^{k-1} s(n-k) \right) v,
\label{eq:xofn}
\end{equation}
assuming the echo state property.  Thus,
\begin{eqnarray}
p_k &=& \langle s(n-k) x(n)\rangle_n \\
&=& \sum_{m=1}^{\infty} W^{m-1} R_{ss}(k-m) v
\label{eq:pk0}
\end{eqnarray}
and
\begin{eqnarray}
C &=& \langle x(n) x(n)^{\top} \rangle_n \\
&=& \sum_{m,m'=1}^{\infty} W^{m-1} vv^{\top} (W^{\top})^{m'-1} R_{ss}(m-m').
\label{eq:C0}
\end{eqnarray}
Substituting Eq.~\ref{eq:WK} into the above equation gives
\begin{eqnarray}
C &=& \frac{1}{2\pi} \sum_{m,m'=1}^{\infty} W^{m-1} vv^{\top} (W^{\top})^{m'-1} \int_{-\pi}^{\pi} S(f) e^{if(m-m')} df \\
&=& \frac{1}{2\pi} \int_{-\pi}^{\pi} S(f) \left(\sum_{m=1}^{\infty} e^{ifm} W^{m-1}\right) vv^{\top} \left(\sum_{m'=1}^{\infty} (W^{\top})^{m'-1}e^{-ifm'}\right) df \\
&=& \frac{1}{2\pi} \int_{-\pi}^{\pi} S(f) \left(\sum_{m=0}^{\infty} e^{ifm} W^{m}\right) vv^{\top} \left(\sum_{m'=0}^{\infty} (W^{\top})^{m'}e^{-ifm'}\right) df \\
&=& \frac{1}{2\pi} \int_{-\pi}^{\pi} S(f) \left(I-e^{if}W\right)^{-1} v v^{\top} \left(1-e^{-if}W^{\top}\right)^{-1} df,
\end{eqnarray}
and using Eq.~\ref{eq:diagonalize},
\begin{eqnarray}
C &=& \frac{1}{2\pi} P \left(\int_{-\pi}^{\pi} S(f) \left(\frac{\omega}{1-e^{if}\vec{d}}\right) \left(\frac{\omega}{1-e^{-if}\vec{d}}\right)^{\top}  df \right)P^{-1}.
\end{eqnarray}
Returning to Eq.~\ref{eq:pk0} and using Eq.~\ref{eq:Rsst}, we have
\begin{eqnarray}
p_k &=& \sum_{m=1}^{\infty} W^{m-1} \left(\sum_{\lambda\in\Lambda} A(\lambda) \lambda^{|k-m|} \right) v \\
&=& \sum_{\lambda\in\Lambda} A(\lambda) \sum_{m=1}^{\infty} W^{m-1} \lambda^{|k-m|} v \\
&=& \begin{cases} \sum_{\lambda\in\Lambda} A(\lambda) \sum_{m=1}^{\infty} W^{m-1} \lambda^{m-k} v & k<1 \\ \sum_{\lambda\in\Lambda} A(\lambda) \left(\sum_{m=1}^{k} W^{m-1} \lambda^{k-m} + \sum_{m=k+1}^{\infty} W^{m-1} \lambda^{m-k} \right) v & k\geq 1 \end{cases} \\
&=& \begin{cases} \sum_{\lambda\in\Lambda} A(\lambda) \lambda^{-k} W^{-1} \left(\sum_{m=1}^{\infty} W^{m} \lambda^m \right) v & k<1 \\ \sum_{\lambda\in\Lambda} A(\lambda) \left(\lambda^k W^{-1} \sum_{m=1}^{k} W^{m} \lambda^{-m} + W^{-1}\lambda^{-k}\sum_{m=k+1}^{\infty} W^{m}\lambda^m \right) v & k\geq 1 \end{cases} \\
&=& \begin{cases} \sum_{\lambda\in\Lambda} A(\lambda) \lambda^{-k} (\lambda^{-1}-W)^{-1} v & k<1 \\ \sum_{\lambda\in\Lambda} A(\lambda) \left( (W^{k}-\lambda^k) (W-\lambda)^{-1}+W^{k}(\lambda^{-1}-W)^{-1}\right) v & k\geq 1 \end{cases}.
\end{eqnarray}
Using Eq.~\ref{eq:diagonalize},
\begin{eqnarray}
p_k &=& P \begin{cases} \sum_{\lambda\in\Lambda} A(\lambda) \lambda^{-k} (\frac{\omega}{\lambda^{-1}-\vec{d}}) & k<1 \\ \sum_{\lambda\in\Lambda} A(\lambda) \text{diag}(\frac{\vec{d}^k-\lambda^k}{\vec{d}-\lambda}+\frac{\vec{d}^k}{\lambda^{-1}-\vec{d}})\omega & k\geq 1 \end{cases}.
\end{eqnarray}
Thus we have
\begin{eqnarray}
PC &=& \sum_{k=0}^{\infty} p_{-k}^{\top} C^{-1} p_{-k} \\
&=& 2\pi \sum_{k=0}^{\infty} \left(\sum_{\lambda\in\Lambda} A(\lambda) \lambda^k (\frac{\omega}{\lambda^{-1}-\vec{d}})\right)^{\top} B^{-1} \left(\sum_{\lambda\in\Lambda} A(\lambda) \lambda^k (\frac{\omega}{\lambda^{-1}-\vec{d}})\right) \\
&=& 2\pi  \sum_{\lambda\in\Lambda} \frac{A(\lambda)A(\lambda')}{1-\lambda\lambda'} (\frac{\omega}{\lambda^{-1}-\vec{d}})^{\top} B^{-1} (\frac{\omega}{(\lambda')^{-1}-\vec{d}}) \\
&=& 2\pi \sum_{i,j} \sum_{\lambda\in\Lambda} \frac{A(\lambda)A(\lambda')}{1-\lambda\lambda'} (\frac{\omega_i}{\lambda^{-1}-d_i}) (B^{-1})_{ij} (\frac{\omega_j}{(\lambda')^{-1}-d_j}) \\
&=& 2\pi \sum_{i,j} \omega_i \left(\sum_{\lambda\in\Lambda} \frac{A(\lambda)A(\lambda')}{1-\lambda\lambda'} \frac{1}{\lambda^{-1}-d_i} \frac{1}{(\lambda')^{-1}-d_j}\right) \left(B^{-1}\right)_{ij} \omega_j,
\end{eqnarray}
which gives the formula in the main text.  Similar manipulations, with the help of Mathematica, give the  more involved formula for $MC$.

\section{Derivation of upper bound for $PC$}

Recall that
\begin{equation}
PC_{\tau} = \frac{\langle s(t+\tau)\hat{s}(t+\tau)\rangle_t^2}{\langle \hat{s}(t)^2\rangle_t}
\end{equation}
and
\begin{equation}
PC = \sum_{\tau=0}^{\infty} PC_{\tau}.
\end{equation}
As our problem setup naturally restricts us to causal linear filters, $PC_{\tau}$ is maximized with $\hat{s}(t+\tau) = \sum_{n=1}^{\infty} s(t-n) k_{\tau}(n)$, with $k_{\tau}(n)$ a Wiener filter.  In particular, suppose that $k_{\tau}(n)$ satisfies the Wiener-Hopf equation:
\begin{equation}
R_{ss}(\tau+t) = \sum_{m=1}^{\infty} R_{ss}(t-m) k_{\tau}(m).
\end{equation}
In matrix form, this reads
\begin{equation}
\begin{pmatrix} R_{ss}(\tau+1) \\ R_{ss}(\tau+2) \\ \vdots \end{pmatrix} = \begin{pmatrix} R_{ss}(0) & R_{ss}(-1) & R_{ss}(-2) & \hdots \\ R_{ss}(1) & R_{ss}(0) & R_{ss}(1) & \hdots \\ \vdots & \vdots & \vdots & \ddots \end{pmatrix} \begin{pmatrix} k_{\tau}(1) \\ k_{\tau}(2) \\ \vdots \end{pmatrix}
\end{equation}
and so
\begin{equation}
 \begin{pmatrix} k_{\tau}(1) \\ k_{\tau}(2) \\ \vdots \end{pmatrix} = \begin{pmatrix} R_{ss}(0) & R_{ss}(-1) & R_{ss}(-2) & \hdots \\ R_{ss}(1) & R_{ss}(0) & R_{ss}(1) & \hdots \\ \vdots & \vdots & \vdots & \ddots \end{pmatrix}^{-1} \begin{pmatrix} R_{ss}(\tau+1) \\ R_{ss}(\tau+2) \\ \vdots \end{pmatrix} .
\end{equation}
For ease of notation, we define $R$ as
\begin{equation}
R := \begin{pmatrix} R_{ss}(0) & R_{ss}(-1) & R_{ss}(-2) & \hdots \\ R_{ss}(1) & R_{ss}(0) & R_{ss}(1) & \hdots \\ \vdots & \vdots & \vdots & \ddots \end{pmatrix}
\end{equation}
and
\begin{equation}
\vec{r}_{\tau} := \begin{pmatrix} R_{ss}(\tau+1) \\ R_{ss}(\tau+2) \\ \vdots \end{pmatrix}
\end{equation}
so in short, $\vec{k}_{\tau} = R^{-1}\vec{r}_{\tau}$.  Then, $\langle s(t+\tau) \hat{s}(t+\tau)\rangle_t = \langle \hat{s}(t)^2\rangle_t$ and so then
\begin{equation}
PC_{\tau} = \langle s(t+\tau)\hat{s}(t+\tau)\rangle_t = \sum_{n=1}^{\infty} R_{ss}(\tau+n) k_{\tau}(n) = \vec{r}_{\tau}^{\top} \vec{k}_{\tau} = \vec{r}_{\tau}^{\top} R^{-1} \vec{r}_{\tau}.
\end{equation}
As these $\vec{k}_{\tau}$ are the causal linear filters that maximize the correlation coefficient between $s(t+\tau)$ and $\hat{s}(t+\tau)$, we have
\begin{equation}
PC \leq \sum_{\tau=0}^{\infty}\vec{r}_{\tau}^{\top} R^{-1} \vec{r}_{\tau}
\end{equation}
for any linear recurrent network.

\end{widetext}

\end{document}

%% file: cmechabbrev.tex
\newcommand{\MeasAlphabet}  {\mathcal{A}}

\newcommand{\MeasSymbol}   { {X} }
\newcommand{\meassymbol}   { {x} }

\newcommand{\CausalState}   { \mathcal{S} }

\newcommand{\causalstate}   { \sigma }
\newcommand{\CausalStateSet}    { \bm{\CausalState} }
\newcommand{\AlternateState}    { \mathcal{R} }

\newcommand{\AlternateStateSet} { \bm{\AlternateState} }

\newcommand{\Prob}      {\Pr} 

\newcommand{\ProcessAlphabet}   {\MeasAlphabet}

\newcommand{\forward}{+}
\newcommand{\reverse}{-}
\newcommand{\forwardreverse}{\pm} 

\newcommand{\FutureCausalState} { {\CausalState}^{\forward} }

\newcommand{\PastCausalState}   { {\CausalState}^{\reverse} }

\newcommand{\lastindex}[2]{
  \edef\tempa{0}
  \edef\tempb{#2}
  \ifx\tempa\tempb
    \edef\tempc{#1}
  \else
    \edef\tempa{0}
    \edef\tempb{#1}
    \ifx\tempa\tempb
      \edef\tempc{#2}
    \else
      \edef\tempc{#1+#2}
    \fi
  \fi
  \tempc
}

\newcommand{\CSjoint}[1][,]{
   \edef\tempa{:}
   \edef\tempb{#1}
   \ifx\tempa\tempb
      \ensuremath{\FutureCausalState\!#1\PastCausalState}
   \else
      \ensuremath{\FutureCausalState#1\PastCausalState}
   \fi
}

\newif\ifpm
\edef\tempa{\forwardreverse}
\edef\tempb{\pm}
\ifx\tempa\tempb
   \pmfalse
\else
   \pmtrue
\fi

%% file: mem_vs_pred.bbl
\begin{thebibliography}{10}

\bibitem{jaeger2001echo}
Herbert Jaeger.
\newblock The ``echo state'' approach to analysing and training recurrent
  neural networks-with an erratum note.
\newblock {\em Bonn, Germany: German National Research Center for Information
  Technology GMD Technical Report}, 148(34):13, 2001.

\bibitem{jaeger2004harnessing}
Herbert Jaeger and Harald Haas.
\newblock Harnessing nonlinearity: Predicting chaotic systems and saving energy
  in wireless communication.
\newblock {\em Science}, 304(5667):78--80, 2004.

\bibitem{lukovsevivcius2009reservoir}
Mantas Luko{\v{s}}evi{\v{c}}ius and Herbert Jaeger.
\newblock Reservoir computing approaches to recurrent neural network training.
\newblock {\em Computer Science Review}, 3(3):127--149, 2009.

\bibitem{Note1}
Memory can be used to estimate the bias of the coin, but nothing else about the
  past provides a guide to the future input.

\bibitem{jaeger2001short}
Herbert Jaeger.
\newblock {\em Short term memory in echo state networks}, volume~5.
\newblock GMD-Forschungszentrum Informationstechnik, 2001.

\bibitem{white2004short}
Olivia~L White, Daniel~D Lee, and Haim Sompolinsky.
\newblock Short-term memory in orthogonal neural networks.
\newblock {\em Physical review letters}, 92(14):148102, 2004.

\bibitem{boedecker2012information}
Joschka Boedecker, Oliver Obst, Joseph~T Lizier, N~Michael Mayer, and Minoru
  Asada.
\newblock Information processing in echo state networks at the edge of chaos.
\newblock {\em Theory in Biosciences}, 131(3):205--213, 2012.

\bibitem{farkavs2016computational}
Igor Farka{\v{s}}, Radom{\'\i}r Bos{\'a}k, and Peter Gergel'.
\newblock Computational analysis of memory capacity in echo state networks.
\newblock {\em Neural Networks}, 83:109--120, 2016.

\bibitem{baranvcok2014memory}
Peter Baran{\v{c}}ok and Igor Farka{\v{s}}.
\newblock Memory capacity of input-driven echo state networks at the edge of
  chaos.
\newblock In {\em International Conference on Artificial Neural Networks},
  pages 41--48. Springer, 2014.

\bibitem{hermans2010memory}
Michiel Hermans and Benjamin Schrauwen.
\newblock Memory in linear recurrent neural networks in continuous time.
\newblock {\em Neural Networks}, 23(3):341--355, 2010.

\bibitem{goudarzi2016memory}
Alireza Goudarzi, Sarah Marzen, Peter Banda, Guy Feldman, Christof Teuscher,
  and Darko Stefanovic.
\newblock Memory and information processing in recurrent neural networks.
\newblock {\em arXiv:1604.06929}, 2016.

\bibitem{ganguli2008memory}
Surya Ganguli, Dongsung Huh, and Haim Sompolinsky.
\newblock Memory traces in dynamical systems.
\newblock {\em Proceedings of the National Academy of Sciences},
  105(48):18970--18975, 2008.

\bibitem{riechers2015pairwise}
Paul~M Riechers, Dowman~P Varn, and James~P Crutchfield.
\newblock Pairwise correlations in layered close-packed structures.
\newblock {\em Acta Crystallographica Section A: Foundations and Advances},
  71(4):423--443, 2015.

\bibitem{fette2005short}
Georg Fette and Julian Eggert.
\newblock Short term memory and pattern matching with simple echo state
  networks.
\newblock {\em Artificial Neural Networks: Biological Inspirations--ICANN
  2005}, pages 13--18, 2005.

\bibitem{toyoizumi2012nearly}
Taro Toyoizumi.
\newblock Nearly extensive sequential memory lifetime achieved by coupled
  nonlinear neurons.
\newblock {\em Neural computation}, 24(10):2678--2699, 2012.

\bibitem{collins2016capacity}
Jasmine Collins, Jascha Sohl-Dickstein, and David Sussillo.
\newblock Capacity and trainability in recurrent neural networks.
\newblock {\em arXiv:1611.09913}, 2016.

\end{thebibliography}
